\documentclass[final,5p]{elsarticle}

\usepackage[T1]{fontenc}
\usepackage[utf8]{inputenc}
\usepackage{amsmath}
\usepackage{booktabs}
\usepackage{array}
\usepackage{tabularx}
\usepackage{graphicx}
\usepackage{xcolor}
\usepackage[colorlinks=true,linkcolor=blue,citecolor=blue,urlcolor=blue]{hyperref}

\graphicspath{{./}}

\journal{Artificial Intelligence in Medicine}

\begin{document}

\begin{frontmatter}

\title{Audit Before You Distill: A Controlled Workflow for Reasoning Transfer into Compact Medical Language Models}

\author[einstein,icti]{Tiago Martinelli\corref{cor1}}
\ead{tiago.martinelli93@gmail.com}

\author[einstein,unesp]{Jo\~ao Paulo Papa}
\ead{joao.papa@unesp.br}

\author[einstein]{Adriano Jos\'e Pereira}
\ead{adriano.pereira@einstein.br}

\cortext[cor1]{Corresponding author.}

\address[einstein]{Einstein Global Advanced Technologies for Equity (E-GATE), Hospital Israelita Albert Einstein, S\~ao Paulo, Brazil}
\address[icti]{Institute of Science and Technology Ita\'u (ICTi), Ita\'u Unibanco, S\~ao Paulo, SP, Brazil}
\address[unesp]{School of Sciences, S\~ao Paulo State University, S\~ao Paulo, Brazil}

\begin{abstract}
Teacher rationales can help a smaller model answer health professions questions, but they can also reinforce errors. We examined when they help and how this evidence can guide training using Brazilian Portuguese questions from medical and multiprofessional residency examinations. First, we measured how a teacher rationale changed the student's probability of selecting the correct answer and compared this effect with rationales drawn from other questions. The complete rationale increased this probability by 21~p.p. when the teacher was correct and decreased it by 14~p.p. when the teacher was wrong. By intervening on the reasoning content, we showed that shuffled rationales did not produce the same pattern. Guided by this result, we trained only with rationales attached to correct teacher answers and compared answer-only training, step-by-step distillation, and our proposed Distill-SFT method. Distill-SFT showed a modest advantage over step-by-step distillation on previously missed questions under reasoning, while using fewer training instances by combining the answer and rationale into a single target. We also investigated catastrophic forgetting using the out-of-domain ENEM dataset. Answer-only training reduced reasoning accuracy by 12~p.p. relative to baseline, compared with approximately 4~p.p. for either rationale-based training method. Together, these findings support a workflow that audits rationale content before distillation, verifies teacher answers before supervision, and trains compact models efficiently for health professions question answering.
\end{abstract}


\end{frontmatter}

\section{Introduction}

Frontier health assistants show how far language models can go, from medical question answering to diagnostic dialogue and health conversations grounded in personal records \cite{singhal2023clinical,tu2025amie,openai2026health}. Bringing this capacity to compact models is attractive in educational and health-informatics settings, where model size and decoding costs constrain deployment. Rationale distillation offers one route: a large teacher generates a rationale for its answer, and that rationale becomes supervision for a smaller student \cite{hsieh2023distilling,shridhar2023reasoning,li2023scotd}.

Previous work has transferred rationales through separate label and explanation tasks \cite{hsieh2023distilling}, direct supervision of smaller models with teacher reasoning traces \cite{shridhar2023reasoning}, multilingual or multi-teacher traces \cite{li2023scotd,tian2024tinyllm}, retrieved knowledge \cite{kang2023kard}, student-oriented trace selection \cite{li2025naturalthoughts,yan2025morsd}, and compressed latent reasoning \cite{shen2025codi}. These approaches improve the way rationales are generated, selected, or encoded. Still, they do not establish by themselves whether transfer is caused by question-relevant content rather than by rationale length, style, or template.

A rationale is not automatically a trustworthy training signal: it may be irrelevant to the question, support an incorrect answer, or constrain the student to a model-specific generation format. These risks are especially consequential outside of English, where compact open-weight students may be weaker, and reasoning behavior may be anchored in another language \cite{payoungkhamdee2024multilingual}. 

Three unresolved issues therefore limit confidence in rationale distillation. First, the benefit of rationale content is often not separated from the mere presence of a long, structured target. Second, teacher correctness may be treated as a filtering detail even though a rationale can transmit an incorrect answer. Third, comparisons based on individual runs can be unstable when accuracy varies substantially across data splits.

Our study asks how reasoning supervision can be diagnosed, selected, and transferred efficiently into a compact model, and why answer verification is essential.

\section{Material and Methods}
To determine whether a rationale carried useful answer-directed content, we designed an exploratory \textit{in silico} study with two linked stages. The work was conducted through a partnership between Hospital Israelita Albert Einstein, a not-for-profit quaternary hospital in São Paulo, Brazil, and São Paulo State University (UNESP), Bauru, Brazil. We distilled a Qwen3-32B teacher into a Qwen3-4B student using Brazilian Portuguese multiple-choice questions from medical and multiprofessional residency examinations. Stage~1 measured the frozen baseline student's answer-choice probabilities while progressively revealing coherent or cross-question-shuffled teacher rationales. A complementary training ablation tested whether the same alignment effect persisted after supervised adaptation. Stage~2 applied the answer-verification rule and compared three training techniques over ten paired data splits: answer-only supervised fine-tuning, step-by-step multi-task distillation, and our proposed single-target distill-SFT. Questions answered correctly by the teacher but incorrectly by the baseline student formed the transfer subset, separating newly acquired answers from answers the student already knew. Statistical evaluation combined two-level bootstrap confidence intervals, paired split-level tests, and within-split McNemar diagnostics.

\paragraph{Data sources and transfer strata} 
Residence exams consist of multiple-choice questions in Brazilian Portuguese. We used multiple-choice examinations because they provide a reproducible test bed for health knowledge and controlled transfer, as in major medical-LLM evaluations. They do not substitute for assessing information gathering, guideline adherence, open-ended communication, or integration into clinical workflows \cite{singhal2023clinical,hager2024clinical,arora2025healthbench,tu2025amie}. 
To probe forgetting of pretrained reasoning behavior outside the adaptation domain, we also used the public Exame Nacional do Ensino Médio (ENEM), Brazil's national secondary-school examination. 
We defined Qwen3-4B before adaptation as the \emph{baseline student}. Each item was assigned to one of four strata according to whether the teacher and baseline student answered it correctly: \textbf{A} (both correct), \textbf{B} (teacher correct, student wrong), \textbf{C} (student correct, teacher wrong), or \textbf{D} (both wrong). Subset~B served as the transfer subset because it isolates questions answered correctly by the teacher but not by the student. 
Aggregate accuracy combines two outcomes---acquiring missing answers on B and preserving answers already known on A and C---that can move in opposite directions and conceal transfer. We therefore report stratum-specific results throughout. 

\paragraph{Models and rationales} 
We used Qwen3-32B as the teacher and Qwen3-4B as the student \cite{qwen3report}. On the analysis pool, the teacher answered $74.9\%$ of items correctly versus $58.8\%$ for the baseline student, leaving a $16.1$-percentage-point transfer gap that motivated distillation. Teacher rationales were generated once and normalized by removing the original thinking delimiters. When a rationale exceeded 250 whitespace-delimited words, preprocessing retained a tail-preserving window aligned to a sentence boundary, thereby preserving its concluding evidence and decision. Only rationales associated with a correct teacher answer were used as training targets. We refer to these as \emph{answer-verified rationales}, because answer verification does not establish that every intermediate statement is valid. The student was adapted with low-rank adaptation (LoRA) \cite{hu2022lora}: rank 32, scaling 32, two epochs, learning rate $10^{-4}$, effective batch size 16, and bfloat16. LoRA initialization was fixed and separated from the ten data-split seeds; the exact seeds are reported in \ref{app:stats}.

\begin{figure*}[!t]
\centering
\includegraphics[width=0.88\textwidth]{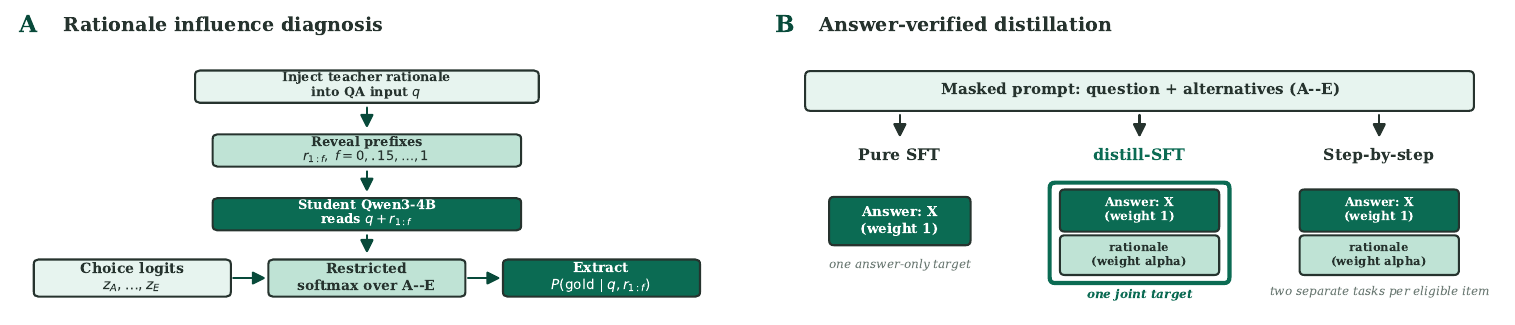}
\caption{\textbf{Study workflow.} Panel A shows how the baseline student read increasing portions of a teacher rationale. After each portion, we measured the probability assigned to the correct answer among options A to E. We compared each rationale with one assigned from another question and separated cases in which the teacher was correct from those in which it was wrong. Panel B shows the training targets. Pure SFT uses the answer only. Distill-SFT places the rationale and answer in one target and gives rationale tokens weight $\alpha$. Step-by-step distillation uses separate answer and rationale tasks, with weight $\alpha$ for the rationale task. Only rationales linked to correct teacher answers were used for training.}
\label{fig:techniques}
\end{figure*}

\paragraph{Study workflow}
Figure~\ref{fig:techniques} shows how the rationale diagnostic in Stage~1 led to the training comparison in Stage~2. The complementary training ablation is summarized in the Results, and the delimiter control is reported in \ref{app:stats}.

In Stage~1, the frozen baseline student read growing prefixes of coherent or cross-question-shuffled teacher rationales. The shuffle preserved length, style, and language while removing relevance to the question, and the trajectories were stratified by whether the teacher was correct. This diagnostic asked whether question-relevant rationale content changed answer-choice probabilities and whether the direction of that change followed the teacher's answer.

\begin{figure*}[!t]
\centering
\includegraphics[width=\textwidth]{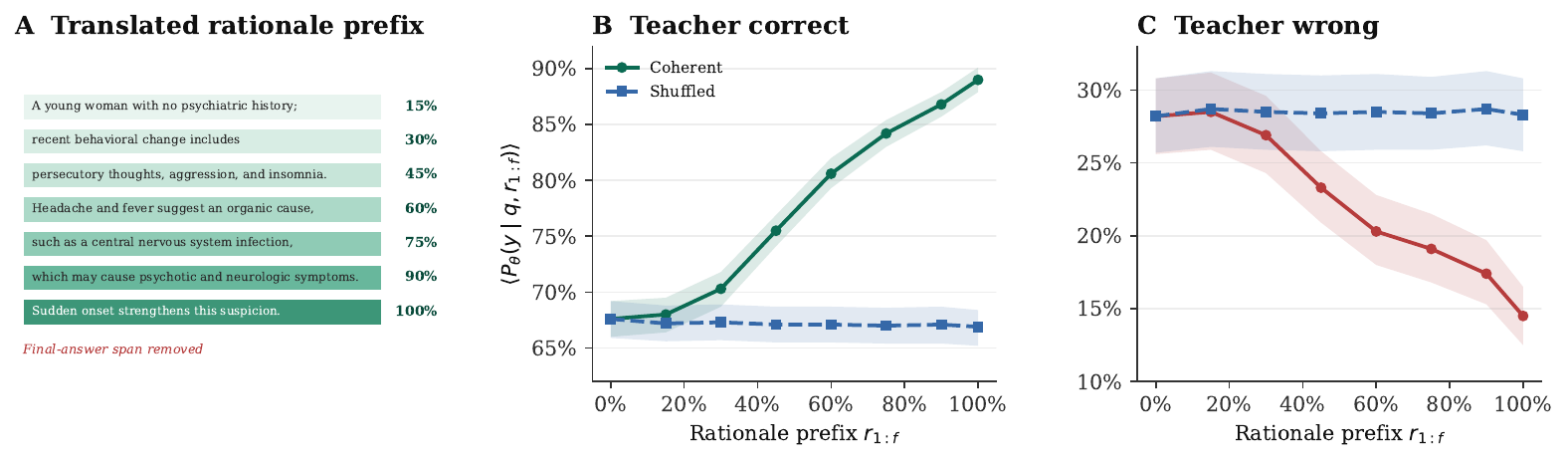}
\caption{\textbf{Rationales move the student toward the teacher's answer.} Panel A shows an English translation of one rationale originally written in Portuguese. The example illustrates how successive parts of the rationale were shown to the student. The curves in panels B and C were calculated using the original Portuguese rationales after removing the explicit final answer. They show the mean probability assigned to the correct answer among the five alternatives when the teacher was correct (B) or wrong (C). Solid lines use the original rationales. Dashed lines use rationales assigned from other questions. Shaded regions show 95\% bootstrap intervals from resampling questions. With the complete rationale, the effects relative to rationales from other questions were $+0.22$ and $-0.14$, respectively, a difference of $0.36$.}
\label{fig:probability}
\end{figure*}

We treated rationale assignment as a controlled intervention: the question, frozen reader, inference prompt, and evaluation rule were held fixed while question--rationale alignment was changed. In the potential-outcomes tradition, causal effects are defined by contrasts between outcomes under alternative interventions on the same treatment variable \cite{holland1986statistics}. The coherent-versus-shuffled contrast therefore estimates the effect of question-relevant rationale content within this protocol; Fig.~\ref{fig:probability} shows the prefix construction and the resulting teacher-correct and teacher-wrong probability trajectories. This protocol-level contrast does not identify the model's internal causal mechanism or establish transportability beyond the evaluated model family, item population, language, and training procedure.

Separately, the complementary training ablation trained matched students with either coherent rationales or cross-question-shuffled rationales while holding the optimization protocol, prompt, and evaluation rule fixed. The models were evaluated on a held-out subset~B test set ($n=164$ per split). This experiment asked whether question--rationale alignment also affected learned transfer; delimiter sensitivity was treated as a secondary control (\ref{app:stats}).

Stage~2 compared three techniques using training data from all four strata. Within each split, A--C were partitioned $85/15$ into training and test items; D was similarly partitioned into training items and a disjoint held-out difficult test. The A--D training partitions were combined in the same model fit, and D was separated only for reporting so that performance on difficult questions remained visible. Train and test items never overlapped. \emph{Pure SFT} supervised the answer only. Our \emph{distill-SFT} used a single completion containing the answer-verified rationale followed by the answer, with rationale tokens weighted by $\alpha$. \emph{Step-by-step} distillation \cite{hsieh2023distilling} created separate answer and rationale tasks. The distill-SFT target retained the Qwen3 thinking delimiters, and reasoning inference began with a Brazilian Portuguese thinking opener. Both rationale-trained techniques were evaluated under direct-answer and reasoning inference.

\paragraph{Rationale-conditioned choice probability}
Before training, the baseline student read the question and growing prefixes $r_{i,f}$ of each stored teacher rationale, with the explicit final-answer span removed. At each checkpoint, we measured how much probability the student assigned to the correct option after reading that fraction of the rationale. We obtained this value by scoring the five answer-letter continuations and normalizing their logits over the restricted A--E set:
\begin{equation}
\widetilde p_{i,f}(a)
=
\frac{\exp z_{i,f,a}}
{\sum_{b\in\{\mathrm{A},\ldots,\mathrm{E}\}}\exp z_{i,f,b}} .
\label{eq:choice-prob}
\end{equation}
For teacher-correct and teacher-wrong groups $g$, the trajectory was the group mean probability of the gold answer,
\begin{equation}
\overline p_{g,c}(f)
=
\frac{1}{|g|}\sum_{i\in g}\widetilde p_{i,f}(y_i),
\qquad c\in\{\mathrm{coherent},\mathrm{shuffled}\}.
\label{eq:mean-choice-prob}
\end{equation}
We reported the full-rationale change from baseline, the coherent-minus-shuffled contrast, and their teacher-correctness interaction. This was a restricted multiple-choice score, not a calibrated full-vocabulary probability.

\paragraph{Loss reweighting} 
We first express the balance between answer and rationale supervision as $\mathcal{L}_{\mathrm{label}}+\alpha\mathcal{L}_{\mathrm{rationale}}$: the label loss trains the final answer, the rationale loss trains the explanation, and $\alpha$ controls their relative contribution. We evaluated $\alpha\in\{1.0,0.3,0.1\}$. Unit weight provided the reference; $0.3$ reduced the contribution of the much longer rationale, and $0.1$ was the lowest weight retained for multi-split evaluation. Values below $0.1$ were not advanced beyond preliminary runs because the rationale objective no longer preserved a consistent reasoning signal. In the separate-task comparator, $\alpha$ scaled the entire rationale task. Our proposed distill-SFT applies this balance within a single completion. Since hundreds of rationale tokens can dominate the loss while the short answer carries the decision, we assigned weight $\alpha$ to rationale tokens and weight $1$ to answer tokens:
\begin{equation}
\begin{aligned}
\mathcal{L}
&= -\frac{\sum_{t} w_t \log p_\theta(y_t\mid y_{<t},x)}
{\sum_t w_t},\\
w_t
&= \begin{cases}
\alpha, & t \in \text{rationale},\\
1, & t \in \text{answer}
\end{cases}.
\end{aligned}
\label{eq:weighted-loss}
\end{equation}
The same $\alpha$ grid therefore varied token weights in our single-target loss and task weights in the comparator. All principal estimates were reconstructed from question-level predictions using a two-level nonparametric bootstrap (data splits resampled, then questions within splits; 10{,}000 replicates) \cite{efron1993bootstrap}, paired $t$ tests across splits with Wilcoxon signed-rank tests as sensitivity analyses, and within-split exact McNemar tests \cite{mcnemar1947}. We considered three principal comparison families: content relevance, technique by inference regime, and rationale loss weight. Delimiter sensitivity was a secondary control reported in \ref{app:stats}. Other pairwise comparisons were exploratory, and the minimum McNemar $p$ across splits was treated as descriptive rather than as a global test. We also monitored generation truncation (maximum $0.45\%$), answer emission (minimum $95.7\%$), and native-English reasoning under the reasoning regime ($0\%$).

\section{Results}
\label{sec:content}

\subsection{Rationales transmit the teacher's answer direction}
The first question was whether rationales carried the teacher's answer into the smaller student, for better or worse. With complete rationales, coherent content increased the mean restricted probability of the gold answer by $0.21$ when the teacher was correct (Fig.~\ref{fig:probability}B) and decreased it by $0.14$ when the teacher was wrong (Fig.~\ref{fig:probability}C). The coherent-versus-shuffled comparison confirmed this directional separation, with a correctness-by-coherence interaction of $0.36$ (95\% cluster-bootstrap CI $0.33$--$0.39$). Coherent rationales thus carried the direction of the teacher's answer: they transferred useful knowledge when the teacher was correct but made an incorrect answer more persuasive when the teacher was wrong. This result motivated the answer-verified supervision used in Stage~2.

The interaction was robust to overt option-wording cues. After jointly excluding rationales with exact normalized gold-option containment or an upper-decile gold-versus-distractor token-overlap margin, the effects retained their direction and the interaction interval remained above zero. This post-hoc sensitivity reused the stored probabilities and could not rule out semantic paraphrase; sample sizes, effect estimates, and confidence intervals are reported in Appendix Table~\ref{tab:lexical}.

A complementary training ablation independently supported the importance of question--rationale alignment: coherent targets transferred better than cross-question-shuffled targets, while a secondary delimiter control found no detectable difference between the two coherent-target formats (\ref{app:stats}).

\subsection{Efficient transfer, selective reasoning, and preserved capability}
\label{sec:technique}

\begin{figure}[t]
\centering
\includegraphics[width=\linewidth]{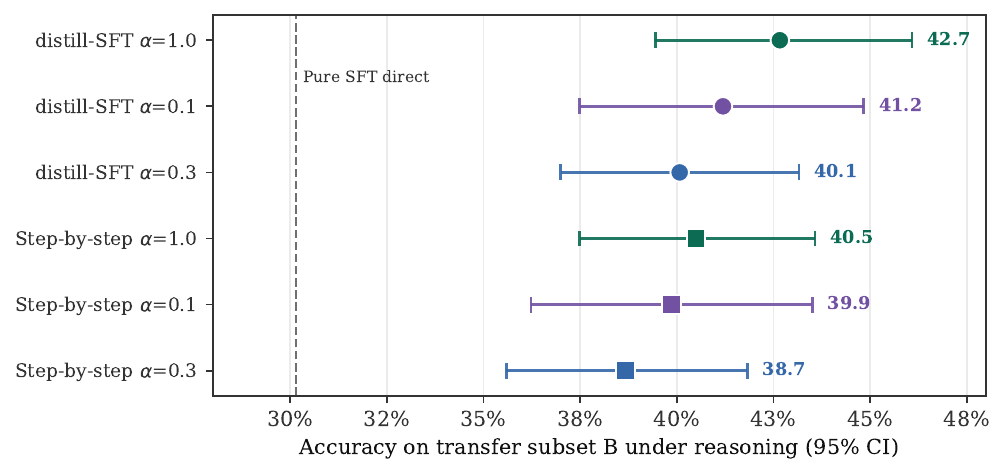}
\caption{\textbf{A localized distill-SFT advantage under reasoning.} Points are ten-split means and lines are 95\% two-level bootstrap intervals on transfer subset~B under reasoning. Color encodes the rationale loss weight $\alpha$ and marker shape the technique. Distill-SFT is numerically ahead of step-by-step at every $\alpha$, with a detected difference at $\alpha=1.0$. The vertical dashed line marks pure-SFT direct answering. Table~\ref{tab:stage2} shows the complementary direct-inference advantage of step-by-step.}
\label{fig:stage2}
\end{figure}

Stage~2 evaluated more than which technique produced the highest aggregate score: could answer-verified rationales recover answers the student lacked, preserve answers it already knew, and do so with a leaner training construction? This balance depended on the inference regime. Table~\ref{tab:stage2} and Fig.~\ref{fig:stage2} compare answer-only SFT, distill-SFT, and multi-task step-by-step distillation. Neither rationale-trained technique was globally dominant. At $\alpha=1.0$, no differences were detected in aggregate ABC accuracy or on difficult subset~D under reasoning. On transfer subset~B, however, distill-SFT gained $2.2$ percentage points over step-by-step under reasoning (95\% paired $t$ CI $0.3$--$4.0$), whereas step-by-step was stronger under direct inference. The technique-by-inference interaction remained positive at the lower rationale weights, supporting an advantage tied to how the student was used rather than uniform superiority. Appendix Table~\ref{tab:paired-effects} reports the paired contrasts, their uncertainty, and the sensitivity tests.

The step-by-step student also required reasoning decoding to reach this performance range. At $\alpha=1.0$, reasoning improved its accuracy on both transfer and difficult items, with gains on subset~B in all ten splits and on subset~D in nine. Its direct mode remained competitive with pure SFT, with no difference detected on either subset, but did not reach the reasoning-enabled transfer regime (Appendix Table~\ref{tab:paired-effects}).

Under matched reasoning inference, distill-SFT used 382 optimizer steps compared with 678 for step-by-step because it learned the rationale and answer in one target rather than duplicating rationale-bearing examples. Their reasoning outputs had similar lengths (303 vs.\ 297 tokens at $\alpha=1.0$), and we detected no difference on difficult subset~D or in the aggregate reasoning comparison. Distill-SFT therefore reached the same broad reasoning-performance range with $44\%$ fewer optimizer updates. Also, no difference was detected between the two rationale-trained methods in the out-of-domain reasoning-retention analysis.

An exploratory frozen-reader audit showed that both techniques generated answer-directed rationales. After removing the explicit answer span, we compared the gold-choice probability induced by each rationale with that induced by a cross-question-shuffled rationale. No between-technique difference was detected in the mean probability shift on B or D. When both generators answered correctly, their rationales increased support for the gold answer; when both answered incorrectly, the effects reversed sign (Appendix Table~\ref{tab:reader-effects}). These measures answer different questions: subset-B accuracy records whether the trained model crossed the final-answer decision boundary, whereas the frozen-reader contrast measures the mean probability shift induced by its rationale. The localized advantage in final decisions was therefore not accompanied by a detected difference in mean rationale influence. This dissociation reinforces the study's central premise: end-task accuracy alone cannot determine which rationales are worth distilling, because their utility is conditional on the item and the receiving student.

Final-answer behavior was also sensitive to how rationale and answer losses were balanced. At full rationale weight, hundreds of rationale tokens dominated the short answer and reduced direct performance on held-out difficult items (subset~D). Down-weighting the rationale improved distill-SFT direct accuracy at both $\alpha=0.3$ and $\alpha=0.1$, without a detected change under reasoning. Step-by-step showed the same pattern: the intermediate weight improved direct accuracy on B and the lowest weight improved it on D, without a detected reasoning improvement in either comparison (Table~\ref{tab:stage2}; paired contrasts in Appendix Table~\ref{tab:paired-effects}). The transfer gain was not achieved by degrading the protected strata: distill-SFT was never significantly worse than step-by-step on A or C and was significantly better on both under reasoning at $\alpha=0.3$ (Appendix Table~\ref{tab:ac}).

\begin{table}[t]
\centering
\small
\setlength{\tabcolsep}{3pt}
\begin{tabular}{@{}lcccccc@{}}
\toprule
Technique & $\alpha$ & Steps & \begin{tabular}[c]{@{}c@{}}B\\dir.\end{tabular} & \begin{tabular}[c]{@{}c@{}}B\\reas.\end{tabular} & \begin{tabular}[c]{@{}c@{}}D\\dir.\end{tabular} & \begin{tabular}[c]{@{}c@{}}D\\reas.\end{tabular} \\
\midrule
Pure SFT & 1.0 & 382 & 0.30 & -- & 0.19 & -- \\
\textbf{distill-SFT} & 1.0 & \textbf{382} & 0.27 & \textbf{0.43} & 0.15 & 0.24 \\
\textbf{distill-SFT} & 0.3 & \textbf{382} & 0.28 & 0.40 & 0.18 & 0.22 \\
\textbf{distill-SFT} & 0.1 & \textbf{382} & 0.29 & 0.41 & 0.18 & \textbf{0.24} \\
Step-by-step & 1.0 & 678 & 0.31 & 0.40 & 0.19 & 0.23 \\
Step-by-step & 0.3 & 678 & 0.33 & 0.39 & 0.21 & 0.23 \\
Step-by-step & 0.1 & 678 & 0.33 & 0.40 & 0.22 & 0.22 \\
\bottomrule
\end{tabular}
\caption{\textbf{Stage~2 accuracy and optimizer steps.} Means across ten data splits. B is the transfer subset and D the held-out difficult subset; ``dir.'' and ``reas.'' denote direct and reasoning inference. A dash marks a regime not evaluated.}
\label{tab:stage2}
\end{table}

\paragraph{Reasoning selectively unlocks difficult-item gains}
\label{sec:cost}

Reasoning did not improve every item, but its balance against the baseline student was favorable. Across the four test strata, the recovery of unknown and difficult items exceeded the erosion of already-known answers by a net $40.5$ correct answers per split. This balance was measured against the baseline student and does not imply global superiority over pure SFT. The benefit was nevertheless selective: distill-SFT was most useful when reasoning-enabled transfer was required, whereas step-by-step was stronger when the deployment objective was direct prediction. Because these experiments used gold-derived strata, they quantify the potential for selective reasoning rather than define a deployable router. Reasoning also increased decoding latency. A prospective gate should therefore predict the \emph{benefit from reasoning}, rather than uncertainty alone \cite{chen2023frugalgpt,ong2024routellm,sprague2024cot,wu2024selectivefiltering}. \ref{app:supp} reports the gain--loss decomposition and the routing and latency sensitivity analyses.

\paragraph{Rationale supervision mitigates forgetting of reasoning behavior}
\label{sec:forgetting}

\begin{figure}[t]
\centering
\includegraphics[width=0.86\linewidth]{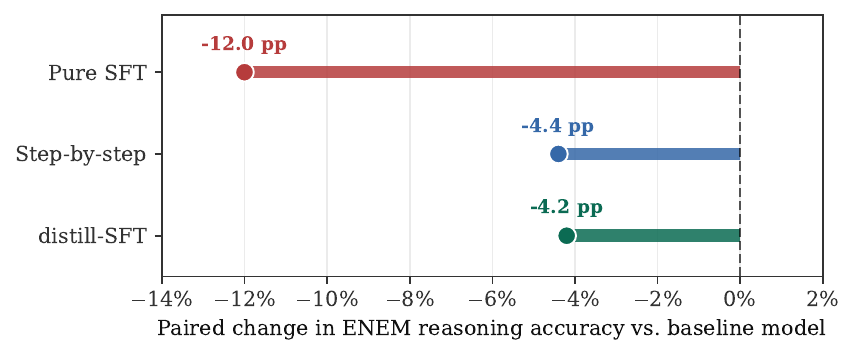}
\caption{\textbf{Rationale supervision mitigates forgetting of pretrained reasoning behavior.} Paired change in out-of-domain ENEM reasoning accuracy relative to the baseline model (ten data splits, $\alpha=1.0$). Answer-only SFT shows the largest degradation; distill-SFT and step-by-step retain more of the pretrained reasoning behavior. The dashed line marks no change from baseline.}
\label{fig:forgetting}
\end{figure}

A student can retain an answer yet forget how to reason through it. We tested this risk on the out-of-domain ENEM examination. Direct-answer accuracy remained within one percentage point of the baseline across all adapted models, but the ability to externalize reasoning did not. The baseline model reached $0.85$ with reasoning, whereas answer-only SFT fell to $0.75$ and produced almost no reasoning text. Both rationale-trained students retained reasoning accuracy near $0.82$; Fig.~\ref{fig:forgetting} reports the paired changes relative to the baseline. The rationale objectives therefore mitigated, but did not eliminate, forgetting of the pretrained reasoning behavior, consistent with rationale-replay findings in language-model adaptation \cite{wu2024radis}.

The rationale objective also transferred the language of reasoning beyond the training domain. On ENEM, which was absent from the adaptation data, the baseline model reasoned predominantly in English, whereas the rationale-trained students externalized their reasoning in Brazilian Portuguese and did so more concisely. The change was therefore not limited to memorizing Portuguese explanations from the medical and multiprofessional residency examinations. It reflected an out-of-domain change in how the compact model expressed its reasoning, without establishing mechanistic faithfulness.

\section{Discussion}

The main result links the rationale diagnostic to the training experiment. A coherent rationale moved the student toward the teacher's answer, whether that answer was right or wrong. A rationale from another question did not have the same effect. We therefore trained only with rationales attached to correct teacher answers. In the comparison across ten data splits, distill-SFT improved transfer on questions the baseline student had missed when the student reasoned. It also used $44\%$ fewer optimizer steps than step-by-step distillation. Step-by-step remained stronger for direct answers. On ENEM, training with rationales reduced the loss of reasoning ability seen after pure SFT, addressing the risk of catastrophic forgetting. Together, these results connect the initial diagnosis to the choice of supervision, transfer, and retention.

Together, the Stage~1 diagnostic and the frozen-reader analysis sharpen this central argument. A localized final-answer gain without a detected difference in mean rationale influence shows why end-task accuracy alone cannot determine which traces are worth distilling. Rationale auditing is therefore not only a safeguard against misleading traces; it can become a selection mechanism that concentrates supervision on rationales that are question-relevant, answer-aligned, and informative for the target student. This student-aware view is consistent with recent rationale-selection methods in general and health-related reasoning distillation \cite{yan2025morsd,song2025rationalequality,yang2026trajectories}.

Distill-SFT placed the answer and rationale signals in one weighted completion, whereas the step-by-step comparator used the separate tasks introduced in prior work. In our generative setting, the step-by-step student required reasoning decoding to reach its best transfer accuracy. When both methods reasoned, their output lengths were similar, distill-SFT had a localized advantage on transfer questions, and single-target construction used substantially fewer optimization steps. Step-by-step was stronger under direct inference, and its direct mode remained competitive with pure SFT. The result is a reduction in training updates for the reasoning-enabled regime, not uniform compute superiority. Reasoning was also unnecessary for every question: gains concentrated on transfer and difficult items, whereas direct answers remained preferable for knowledge already accessible to the student. These findings support a compact student with direct reasoning and behavioral behaviors and motivate a prospective gate that invokes reasoning only when it is likely to correct a direct response, consistent with selective-filtering approaches \cite{wu2024selectivefiltering,sprague2024cot}.

The design had three strengths. We changed the rationale assigned to a question while keeping the student, prompt, and answer scoring fixed. We then tested the role of this alignment again during training. We also compared the methods on the same questions across ten data splits and reported transfer separately from answers the student already knew. Finally, ENEM let us assess whether the student retained its reasoning behavior outside the residency examination domain. These choices make the gains and losses easier to interpret without treating a difference that was not detected as proof that two methods are equivalent.

\subsection{Limitations}
\label{sec:limitations}

The findings should be interpreted within the scope of the evaluated protocol. Answer verification applied only to the final option, so a correct answer could still rest on false, incomplete, or unsafe intermediate statements. The workflow is a first filter, not a factuality or clinical-safety assessment. The frozen-reader audit measured answer-directed influence rather than truth, completeness, or clinical safety; independent factual and clinical evaluation of generated rationales remains future work. Automatic answer extraction from unconstrained model outputs is a recognized source of evaluation error \cite{xfinder}. In our independent parsing audit, the estimated teacher accuracy changed by less than one percentage point, indicating that the main estimate was stable, although residual errors may affect a small fraction of teacher-correct and teacher-wrong assignments (details in \ref{app:supp}). The probability diagnostic was conditional on one reader, prompt, and restricted five-option scoring scheme. More broadly, the study covered one model family, one language, and multiple-choice examinations; selecting the keyed option does not establish open-ended clinical competence \cite{wang2025leastwrong}. The ten data splits quantified sensitivity within the same item pool rather than across independent populations, and the gold-derived strata cannot serve as a deployed reasoning router. The study therefore establishes a controlled methodological foundation in medical-residency examinations, while broader prospective validation remains necessary before clinical deployment.

\section{Conclusion}

In Brazilian Portuguese residency examinations, question-relevant rationales moved a compact student toward the teacher's answer, including when that answer was wrong. This finding supports answer verification as a first filter for rationale supervision. Under paired evaluation, single-target distill-SFT used $44\%$ fewer optimizer updates than step-by-step distillation and achieved higher accuracy on the target transfer subset under reasoning, whereas step-by-step was stronger under direct inference. Rationale supervision also mitigated forgetting of pretrained reasoning behavior outside the training domain. Together, these findings establish an auditable workflow for evaluating rationale supervision and its consequences when adapting compact models for health applications. Prospective evaluation on clinical tasks is needed to determine whether these benefits extend beyond examination performance.

\section*{Ethics statement}
Ethics approval and informed consent were not required because the study involved no human participants, patient data, clinical records, or identifiable personal information. All experiments used publicly available examination questions and model-generated outputs.

\section*{Data availability}
The study repository contains the question-level study data, model-generated teacher rationales, canonical seed definitions, aggregate analysis tables, and code needed to audit the reported analyses and construct comparable 85/15 seeded partitions. The original examining bodies retain all rights to the examination content; provenance and use restrictions are documented with the released data.

\section*{Code availability}
Training, inference, masking self-tests, loss-weighting tests, and statistical analysis code are available at \url{https://github.com/martinellitiago/cot-distillation-medical-qa}. The versioned archive is available on Zenodo at \url{https://doi.org/10.5281/zenodo.22929248}.

\section*{CRediT authorship contribution statement}
Tiago Martinelli: Conceptualization, Methodology, Software, Formal analysis, Investigation, Data curation, Visualization, Writing--original draft, Project administration. João Paulo Papa: Conceptualization, Validation, Supervision, Writing--review and editing. Adriano José Pereira: Conceptualization, Validation, Resources, Supervision, Funding acquisition, Writing--review and editing.

\section*{Declaration of AI-assisted technologies in the manuscript preparation process}
During the preparation of this work, the authors used OpenAI Codex for language editing and Anthropic Claude Code to assist with the development and checking of analysis scripts. The authors reviewed and validated all text and code and take full responsibility for the content of the publication.

\section*{Declaration of competing interest}
The authors declare that they have no known competing financial interests or personal relationships that could have appeared to influence the work reported in this paper.

\section*{Funding}
This work was supported by capital expenditure (CAPEX) funding from Hospital Israelita Albert Einstein, a not-for-profit private hospital in São Paulo, Brazil, which coordinated the study. The funding provided a postdoctoral fellowship for TM and supported an academic visit by JPP as a visiting professor from São Paulo State University (UNESP), Bauru, Brazil.

\section*{Acknowledgements}
The authors thank the Digital Transformation Office (DTO) and the leadership of Einstein Global Advanced Technologies for Equity for their support throughout this study. We also acknowledge the Institute of Science and Technology Ita\'u (ICTi) for providing the computational infrastructure used during the initial phase of this research. We also thank Cesar Gomes Miguel (Data Scientist Specialist), Gabriela Chiuffa Tunes (Manager of Advanced Analytics and Multiomics), José Adenaldo Santos Bittencourt Júnior (Manager of High Performance Computing), and Edson Amaro Junior (Director of Global Data and Advanced Technologies) for their endorsement and assistance.

\appendix
\numberwithin{equation}{section}
\renewcommand{\theequation}{\Alph{section}.\arabic{equation}}

\section{Training and analysis}
\label{app:stats}

\paragraph{Token-weighted distill-SFT loss} 
In distill-SFT, rationale tokens receive weight $\alpha$ and answer tokens receive weight $1$ in Eq.~\ref{eq:weighted-loss}. When no rationale is available, the target contains only the answer. We used $\alpha\in\{1.0,0.3,0.1\}$. For illustration, with 50 rationale tokens per answer token, these weights give the answer approximately $2\%$, $6\%$, and $17\%$ of the total token weight, respectively. These percentages describe the weights assigned to tokens; the actual contribution to the loss also depends on each token's prediction error. In step-by-step distillation, $\alpha$ weights the separate rationale task, while the answer task keeps weight $1$. We evaluated both methods at all three weights and included pure SFT, giving seven configurations across ten data splits (70 runs).

\paragraph{Cross-question-shuffled control} 
Over the eligible teacher-correct items we draw, with a split-specific generator, a permutation $\pi$ with no fixed point ($\pi(i)\neq i$ for all $i$), so each item receives another item's rationale of comparable length, style, and language while its relevance to the question is destroyed. A hard assertion verifies zero self-assignments; across the ten canonical data splits none occurred.

\paragraph{Complementary training ablation and delimiter sensitivity}
Matched students were trained with coherent or cross-question-shuffled rationale targets under the same optimization protocol and evaluated on a held-out subset~B test set ($n=164$ per split). Coherent rationales yielded mean accuracies of $0.390$ with Qwen3 thinking delimiters and $0.391$ as ordinary body text, whereas the corresponding shuffled-rationale accuracies were $0.237$ and $0.285$. No difference was detected between the two coherent-target formats (paired $p=0.91$). Qwen3 supports thinking and non-thinking modes \cite{qwen3report}; this secondary control does not establish equivalence between formats or imply that delimiters are generally irrelevant. It shows only that the delimiter provided no detectable benefit to coherent supervised transfer in this protocol.

\paragraph{Two-level bootstrap} 
We resample data splits with replacement and, within each chosen split, resample questions with replacement, over $10^4$ replicates; the reported interval is the $2.5$/$97.5$ percentile of the resulting distribution of the across-split mean.

\paragraph{Paired tests and effect reporting} 
The primary inferential comparison is a paired $t$ test on the ten split-level accuracies, with the Wilcoxon signed-rank test as a sensitivity analysis. Within-split exact McNemar tests on discordant question pairs are item-level diagnostics; the minimum McNemar $p$ across splits is descriptive and is never interpreted as a single global test. The manuscript focuses on three principal comparison families: content relevance, technique by inference regime, and rationale loss weight; delimiter sensitivity is treated as a secondary control. Other pairwise comparisons are treated as exploratory. Non-significant differences are reported as \emph{not detected}, never as equivalence, because no equivalence margin was prespecified. Effects are given in percentage points and in questions gained or lost per split. Table~\ref{tab:paired-effects} consolidates the Stage~2 contrasts discussed in the main text. Its confidence intervals are paired $t$ intervals computed from the ten split-level differences with nine degrees of freedom, distinct from the two-level bootstrap intervals for accuracy summaries. Tests and intervals are two-sided and unadjusted for multiplicity.

\begin{table*}[!t]
\centering
\small
\setlength{\tabcolsep}{5pt}
\begin{tabularx}{\textwidth}{@{}Xllccc@{}}
\toprule
Contrast & Set & Inference & Difference, p.p. (95\% CI) & $t$-$p$ & Wilcoxon $p$ \\
\midrule
\multicolumn{6}{@{}l}{\textit{Technique and inference regime at $\alpha=1.0$, unless specified}} \\
distill-SFT minus step-by-step & B & reasoning & $+2.17$ [$+0.33$, $+4.00$] & 0.025 & 0.035 \\
distill-SFT minus step-by-step & B & direct & $-3.57$ [$-4.76$, $-2.38$] & $<0.001$ & 0.002 \\
distill-SFT minus step-by-step & ABC & reasoning & $+1.04$ [$-0.40$, $+2.48$] & 0.137 & 0.221 \\
distill-SFT minus step-by-step & D & reasoning & $+0.59$ [$-1.59$, $+2.78$] & 0.553 & 0.711 \\
Technique-by-inference interaction & B & both & $+5.73$ [$+3.00$, $+8.47$] & 0.001 & 0.004 \\
Interaction at $\alpha=0.3$ & B & both & $+6.22$ [$+2.21$, $+10.24$] & 0.007 & 0.014 \\
Interaction at $\alpha=0.1$ & B & both & $+4.62$ [$+0.46$, $+8.77$] & 0.033 & 0.025 \\
Step-by-step: reasoning minus direct & B & both & $+9.93$ [$+7.07$, $+12.79$] & $<0.001$ & 0.002 \\
Step-by-step: reasoning minus direct & D & both & $+3.71$ [$+1.86$, $+5.57$] & 0.001 & 0.006 \\
Step-by-step minus pure SFT & B & direct & $+0.42$ [$-2.13$, $+2.97$] & 0.719 & 0.691 \\
Step-by-step minus pure SFT & D & direct & $+0.30$ [$-1.93$, $+2.53$] & 0.770 & 0.641 \\
\midrule
\multicolumn{6}{@{}l}{\textit{Rationale-weight contrasts: reduced weight minus $\alpha=1.0$}} \\
distill-SFT, $\alpha=0.3$ & D & direct & $+2.57$ [$+0.27$, $+4.88$] & 0.032 & 0.043 \\
distill-SFT, $\alpha=0.1$ & D & direct & $+2.82$ [$+0.75$, $+4.89$] & 0.013 & 0.012 \\
distill-SFT, $\alpha=0.3$ & D & reasoning & $-1.34$ [$-3.44$, $+0.77$] & 0.185 & 0.285 \\
distill-SFT, $\alpha=0.1$ & D & reasoning & $+0.35$ [$-1.71$, $+2.41$] & 0.712 & 0.900 \\
Step-by-step, $\alpha=0.3$ & B & direct & $+2.17$ [$+0.22$, $+4.12$] & 0.033 & 0.043 \\
Step-by-step, $\alpha=0.1$ & D & direct & $+2.33$ [$+0.59$, $+4.06$] & 0.014 & 0.010 \\
Step-by-step, $\alpha=0.3$ & B & reasoning & $-1.82$ [$-4.78$, $+1.15$] & 0.199 & 0.227 \\
Step-by-step, $\alpha=0.1$ & D & reasoning & $-1.09$ [$-2.76$, $+0.58$] & 0.174 & 0.219 \\
\bottomrule
\end{tabularx}
\caption{\textbf{Paired training comparisons.} Differences are in p.p. across ten paired splits, with paired $t$ intervals. The interaction is the reasoning contrast minus the direct contrast for distill-SFT versus step-by-step. Tests are two-sided and unadjusted; Wilcoxon tests provide a sensitivity check.}
\label{tab:paired-effects}
\end{table*}

\paragraph{Confounder panel} 
We monitored three generation-related confounders: truncation at the generation budget (maximum $0.45\%$), answer emission (minimum $95.7\%$), and native-English reasoning under the reasoning regime ($0\%$). None materially affected the principal comparisons.

\paragraph{Reproducibility}
We used a single NVIDIA H100 80-GB GPU with bfloat16 precision. Inference used greedy decoding. The ten data-split seeds were $8,12,17,23,25,31,37,44,52,$ and $61$; LoRA initialization used seed $3407$. Latency measurements covered generation only, with GPU synchronization before and after each call and the first call after each restart excluded. Software versions and execution settings are documented at \url{https://github.com/martinellitiago/cot-distillation-medical-qa}.

\section{Additional analyses}
\label{app:probability}
\label{app:supp}

The teacher-answer check and the initial probability diagnostic used all 4{,}260 question records. These records contained 3{,}792 distinct normalized stems because some stems appeared more than once. All records were kept in the initial diagnostic; records sharing a stem were sampled together in the bootstrap. The lexical sensitivity checks then repeated the comparisons after excluding records with option-wording cues.

\subsection{Answer extraction}
We checked whether the rule used to extract the teacher's final answer changed its measured accuracy. Extraction errors in free-text responses are a known source of evaluation error \cite{xfinder}. We applied a second parser that prioritized explicit final-answer phrases, such as \texttt{Resposta correta: X}, and the bold \texttt{X)} format. The two parsers returned different answers for 77 of 4{,}260 records ($1.81\%$). Against the official answer key, only the original parser was correct in 44 cases, only the second parser was correct in 18, and both were wrong in 15. Estimated teacher accuracy changed from $74.9\%$ to $74.3\%$. All training comparisons retained the original labels and item partitions. Neither parser was treated as a reference standard for ambiguous outputs.

\subsection{Rationale probabilities}

\paragraph{Answer scoring} 
For each question and rationale fraction, the reader produces logits for the five answer-letter continuations. Equation~\ref{eq:choice-prob} applies a softmax only across these five candidates. The explicit final-answer span is removed from the rationale before scoring to prevent label leakage. Prefixes are evaluated at fractions $0$, $0.15$, $0.30$, $0.45$, $0.60$, $0.75$, $0.90$, and $1.0$. Coherent rationales are compared with a fixed cross-question-shuffled control, and all reported effects remain on the probability scale.

\paragraph{Probability contrasts} 
We made two comparisons, separately for questions the teacher answered correctly and incorrectly. Let $g$ denote either group and $\overline p_{g,c}(f)$ the mean probability of the correct answer in that group, where $c$ identifies the rationale condition and $f$ is the fraction shown to the student. Thus $f=0$ means no rationale and $f=1$ means the complete rationale after removal of the explicit final answer.

First, we measured the change after adding the question's own rationale:
\begin{equation}
d_g=\overline p_{g,\mathrm{coherent}}(1)-\overline p_{g,\mathrm{coherent}}(0).
\label{eq:app-rationale-change}
\end{equation}
Second, we compared that rationale with one assigned from another question:
\begin{equation}
C_g=\overline p_{g,\mathrm{coherent}}(1)-\overline p_{g,\mathrm{shuffled}}(1).
\label{eq:app-rationale-control}
\end{equation}
Positive values indicate greater support for the correct answer; negative values indicate less support. The interaction compares this second effect between the two teacher groups:
\begin{equation}
I=C_{\mathrm{correct}}-C_{\mathrm{wrong}}.
\label{eq:app-rationale-interaction}
\end{equation}
For example, $C_{\mathrm{correct}}=+0.22$ and $C_{\mathrm{wrong}}=-0.14$ give $I=0.36$, or 36~p.p. This is a difference in probability shifts, not an accuracy gain.

Confidence intervals were calculated by resampling the stem groups described above.

\paragraph{Lexical-overlap sensitivity} 
We tested whether the probability shifts could be explained by rationales repeating the wording of the correct option. After normalizing case, accents, and punctuation, we flagged rationales containing the full correct option, provided it had at least two content words or 12 characters. We also measured the fraction of each option's content words present in the rationale, excluding Portuguese stopwords and words from the question stem. When removing stem words left no content words, we used the option's original content words. The overlap margin was the coverage of the correct option minus the highest coverage among the incorrect options. We flagged values at or above its 90th percentile.

We repeated the comparisons after excluding either type of flagged record separately and then both together (Table~\ref{tab:lexical}). This post-hoc analysis reused the saved probabilities and stem groups; the models were not run again. In each analysis, bootstrap samples included only the remaining records and their stem groups. After both exclusions, 2{,}920 records in 2{,}661 groups remained. The effects were $+0.18$ when the teacher was correct and $-0.12$ when it was wrong, with an interaction of $0.30$ (95\% CI $0.27$ to $0.33$). The same direction therefore persisted after removing these wording cues, although this check cannot rule out paraphrases of the correct option.

\begin{table*}[!t]
\centering
\small
\setlength{\tabcolsep}{5pt}
\begin{tabular}{@{}lrrrrr@{}}
\toprule
Sensitivity set & Records & Stem groups & $C_{\mathrm{correct}}$ & $C_{\mathrm{wrong}}$ & $I$ (95\% CI) \\
\midrule
All records & 4,260 & 3,792 & $+0.22$ & $-0.14$ & $+0.36$ [$+0.33$, $+0.39$] \\
Exclude exact option wording & 3,145 & 2,850 & $+0.18$ & $-0.12$ & $+0.30$ [$+0.27$, $+0.33$] \\
Exclude high overlap margin & 3,825 & 3,432 & $+0.21$ & $-0.14$ & $+0.35$ [$+0.32$, $+0.38$] \\
Exclude both types of cue & 2,920 & 2,661 & $+0.18$ & $-0.12$ & $+0.30$ [$+0.27$, $+0.33$] \\
\bottomrule
\end{tabular}
\caption{\textbf{Lexical sensitivity.} Records count question entries; stem groups collect repeated question stems. Effects follow Eqs.~\ref{eq:app-rationale-control} and \ref{eq:app-rationale-interaction}, with bootstrap intervals over the remaining groups.}
\label{tab:lexical}
\end{table*}

\begin{table*}[!t]
\centering
\small
\setlength{\tabcolsep}{5pt}
\begin{tabular}{@{}llrccccc@{}}
\toprule
Set & Generator outcomes & $n$ & Distill & Step & Difference (95\% CI) & $t$-$p$ & Wilcoxon $p$ \\
\midrule
B & All & 1,430 & $+0.21$ & $+0.19$ & $+0.02$ [$-0.02$, $+0.05$] & 0.183 & 0.232 \\
B & Both correct & 417 & $+0.59$ & $+0.56$ & $+0.03$ [$-0.02$, $+0.09$] & 0.120 & 0.193 \\
B & Both wrong & 658 & $-0.11$ & $-0.09$ & $-0.02$ [$-0.06$, $+0.02$] & 0.101 & 0.065 \\
D & All & 2,020 & $+0.08$ & $+0.08$ & $-0.01$ [$-0.04$, $+0.03$] & 0.630 & 0.770 \\
D & Both correct & 259 & $+0.64$ & $+0.62$ & $+0.02$ [$-0.07$, $+0.10$] & 0.553 & 0.322 \\
D & Both wrong & 1,333 & $-0.11$ & $-0.10$ & $-0.01$ [$-0.04$, $+0.01$] & 0.132 & 0.049 \\
\bottomrule
\end{tabular}
\caption{\textbf{Generated rationales.} Distill and Step show the change in correct-answer probability with the question's own rationale versus another question's rationale. Differences are distill-SFT minus step-by-step, with bootstrap intervals. $n$ counts paired evaluations across splits. Outcome groups are descriptive; tests are unadjusted.}
\label{tab:reader-effects}
\end{table*}

\paragraph{Generated rationales}
We tested whether rationales produced by the trained models helped the baseline Qwen3-4B choose the correct answer. We used outputs from distill-SFT and step-by-step at $\alpha=1.0$ on the same ten test splits of B and D. After removing the explicit final answer, we measured the baseline model's answer probabilities with no rationale, with the question's own rationale, and with one from another question. Its parameters remained unchanged throughout. We compared the last two conditions using the same paired tests and bootstrap procedure (Table~\ref{tab:reader-effects}). We also summarized cases in which both trained models answered correctly or both answered incorrectly; these groups were defined after generation and are descriptive. This test measures answer influence within the Qwen family. It does not verify the factual content of the explanation.

\subsection{Transfer and retention}
Table~\ref{tab:ac} compares accuracy on A and C, the questions the baseline student already answered correctly. No lower accuracy was detected for distill-SFT than for step-by-step in these strata. At $\alpha=0.3$, distill-SFT had higher accuracy on both under reasoning.

\begin{table}[!htbp]
\centering
\small
\setlength{\tabcolsep}{4pt}
\begin{tabular}{@{}llccccc@{}}
\toprule
Sub. & Regime & $\alpha$ & Step & Distill & $\Delta$ (p.p.) & $t$-$p$ \\
\midrule
A & direct    & 1.0 & 93.1 & 93.9 & $+0.8$ & 0.20 \\
A & direct    & 0.3 & 93.4 & 93.4 & $+0.0$ & 0.95 \\
A & direct    & 0.1 & 92.7 & 93.1 & $+0.4$ & 0.46 \\
A & reasoning & 1.0 & 82.0 & 82.7 & $+0.7$ & 0.40 \\
\textbf{A} & \textbf{reasoning} & \textbf{0.3} & 80.0 & 82.2 & $\mathbf{+2.2}$ & \textbf{0.02} \\
A & reasoning & 0.1 & 81.9 & 82.5 & $+0.6$ & 0.55 \\
C & direct    & 1.0 & 79.6 & 80.7 & $+1.1$ & 0.56 \\
C & direct    & 0.3 & 78.0 & 80.7 & $+2.8$ & 0.14 \\
C & direct    & 0.1 & 77.8 & 80.9 & $+3.2$ & 0.18 \\
C & reasoning & 1.0 & 61.1 & 61.1 & $+0.0$ & 1.00 \\
\textbf{C} & \textbf{reasoning} & \textbf{0.3} & 60.7 & 67.2 & $\mathbf{+6.5}$ & \textbf{0.006} \\
C & reasoning & 0.1 & 63.2 & 62.6 & $-0.6$ & 0.76 \\
\bottomrule
\end{tabular}
\caption{\textbf{Retention of previously correct answers.} Mean accuracy on subsets A and C in percent across ten splits. Differences are distill-SFT minus step-by-step in p.p., with paired $t$ tests. Bold rows indicate $p<0.05$.}
\label{tab:ac}
\end{table}

\paragraph{Selective reasoning}
With reasoning at $\alpha=1.0$, distill-SFT gained a mean of $61.0$ correct answers on B and $47.9$ on D relative to the baseline student, but lost $47.4$ on A and $21.0$ on C. The net gain was $40.5$ correct answers per split. Against pure-SFT direct answering, it gained $17.9$ on B and $9.3$ on D but lost $26.2$ on A and $7.4$ on C. Reasoning therefore helped on B and D without improving overall accuracy over pure SFT. Table~\ref{tab:routing} compares accuracy and latency when reasoning is used on all items or selected items. Selection by subset or by an oracle requires knowing which answers are correct, so these results describe potential gains rather than a tested deployment rule.

\begin{table}[!htbp]
\centering
\small
\setlength{\tabcolsep}{4pt}
\begin{tabular}{@{}llcc@{}}
\toprule
Set & Offline policy & Acc. & Latency (\%) \\
\midrule
ABC  & answer all directly & 0.72 & 1 \\
ABC  & route subset B to reasoning & 0.77 & 31 \\
ABC  & reason on all items & 0.68 & 100 \\
ABC  & per-question oracle & 0.81 & 10 \\
\midrule
D & answer all directly & 0.15 & 1 \\
D & reason on all items & 0.24 & 100 \\
D & per-question oracle & 0.31 & 17 \\
\bottomrule
\end{tabular}
\caption{\textbf{Selective reasoning.} Distill-SFT at $\alpha=1.0$ across ten splits. Latency is relative to reasoning on every item. Subset and oracle policies use known answer outcomes and cannot be applied prospectively as tested.}
\label{tab:routing}
\end{table}

\begingroup
\renewcommand{\url}[1]{}
\interlinepenalty=10000
\bibliographystyle{elsarticle-num}
\bibliography{main}
\endgroup

\end{document}